\documentclass[conference]{ieeeconf}

\usepackage{graphics} 
\usepackage{graphicx}
\usepackage{cite}

\title{\LARGE \bf
Embracing Contact: Pushing Multiple Objects with Robot's Forearm
}

\author{Akansel Cosgun, Luke Ditria, Shayne D'Lima and Tom Drummond\\
ARC Centre for Excellence for Robot Vision\\
Monash University, Australia}

\begin{document}

\maketitle
\thispagestyle{empty}
\pagestyle{empty}

\begin{abstract}
Grasping is the dominant approach for robot manipulation, but only a single object can be grasped at a time. Nonprehensile manipulation offers richer set of interactions, however state-of-the-art is limited to using the end-effector only. We propose using a robot link (forearm) to push multiple objects at once. In a simulated task where the robot's task is to sort two kinds of objects into their respective goal regions, we show that a greedy strategy that uses a combination of forearm pushes and pick and place operations reduces task completion time by \%28 compared to picking and placing each object individually.
\end{abstract}

\section{Introduction}
\label{sec:introduction}

The majority of the robot manipulation research is concerned with the grasping problem~\cite{bohg2013data}. Pick and place remains the most popular method for robotic manipulation because once a rigid object is grasped, it can be considered the part of the kinematic chain and that simplifies the placement problem. A limitation of pick and place, however, is that only a single object can be grasped at a time.

Non-prehensile manipulation offers alternative strategies to object manipulation~\cite{lynch1999dynamic,barry2013manipulation}. Robotic pushing can be used to facilitate object segmentation~\cite{hermans2012guided}, placement~\cite{cosgun2011push}, grasping~\cite{dogar2011framework}, setting up a table~\cite{scholz2010combining}, handling book-like objects~\cite{cosgun2015stacking} and dirt rearrangement~\cite{elliott2018robotic}. A comprehensive survey on robotic pushing can be found in~\cite{stuber2019let}. Literature in this field is mostly limited to contact with the end-effector as robotics research, and motion planning in particular, considers any contact with the body a `collision'. We propose allowing contact between the robot links with objects, which enables new strategies for tabletop object manipulation. In particular, we utilize the forearm link to manipulate multiple objects at once. Previous works have considered haptic sensing with the forearm~\cite{bhattacharjee2012haptic} and exploiting contact to reduce positional uncertainty \cite{pall2018contingent}; however to our knowledge no prior work has considered deliberately manipulating objects with the forearm of a serial manipulator.

In simulation experiments, we show that the a greedy strategy using a mix of pushing and pick$\&$place actions sorts tabletop objects more efficiently than pick$\&$place only. The approach is also implemented on a UR5 robot.

\begin{figure}[ht!]
\centering
\includegraphics[clip, trim={5cm 4cm 10.5cm 4cm}, width=0.22\textwidth]{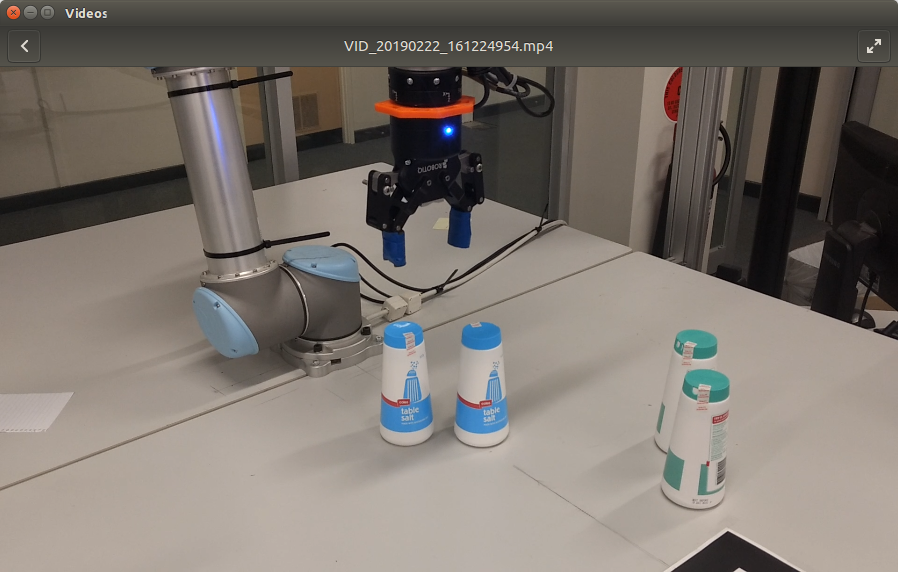}
\includegraphics[clip, trim={5cm 4cm 10.5cm 4cm}, width=0.22\textwidth]{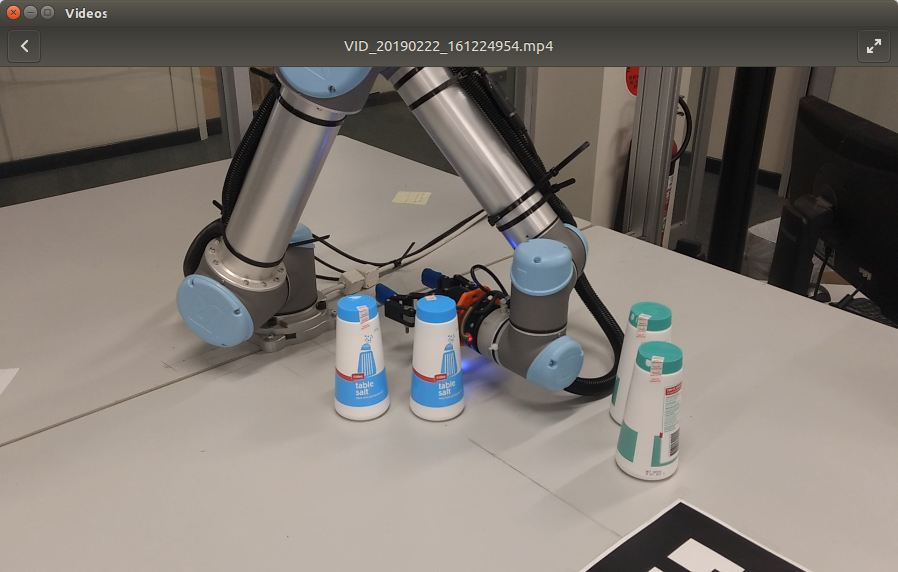} \\ 
\vspace{0.08cm}
\includegraphics[clip, trim={5cm 4cm 10.5cm 4cm}, width=0.22\textwidth]{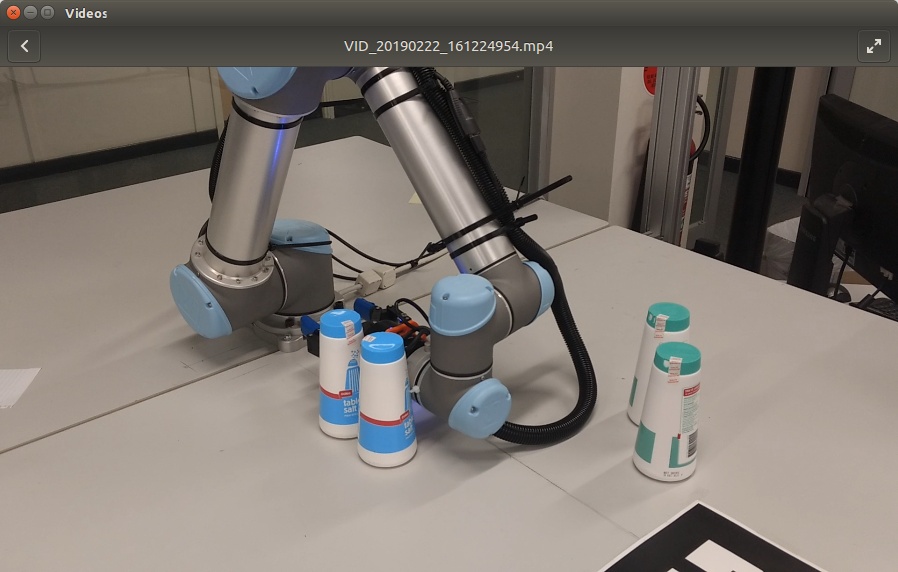}
\includegraphics[clip, trim={5cm 4cm 10.5cm 4cm}, width=0.22\textwidth]{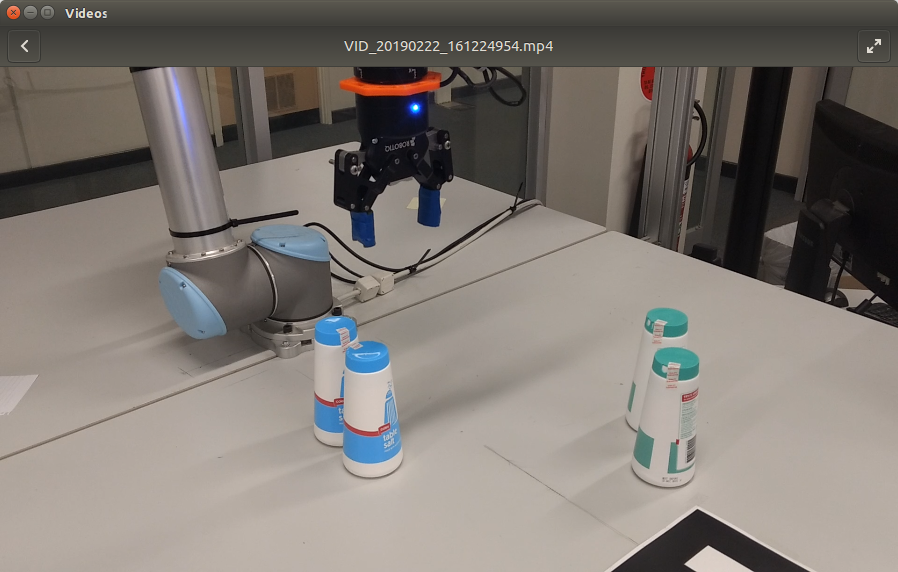}
\caption{Steps of a forearm-push action is shown. Pushing multiple objects together at once can increase the efficiency of certain tasks.}
\vspace{-0.3cm}
\label{fig:intro}
\end{figure}

\section{Pushing with Forearm}
\label{sec:approach}

We define a push action by two points: the desired starting and ending positions. We constrain the available pushing trajectories to ones where the end-effector moves on a virtual line parallel to the table surface. The push is executed in three steps: approach, push and retract. Before executing the push action, we first find the actual push start position and check the kinematic feasibility of each of the three motions. If the forearm would be in collision with the objects in the desired push start position, we search for a collision-free start position in the reverse direction of the push. The robot is position-controlled during the push via intermediate trajectory points with fixed end-effector speed. During the push motion, the end effector pose is constrained such that it has a fixed height with respect to the table and the orientation is kept fixed, parallel to the table. In order to estimate where the objects will end up given a proposed push action we use Box2D, a dynamic 2D simulator. Simple 2D physics is a sufficient model to predict the outcome of a push for our purposes but learning-based methods would be more suitable if objects with different geometries are used~\cite{finn2017deep}.

\section{Experiments}

\subsection{Sorting Task}
The task of the robot is to sort two kinds of cylindrical objects (blue and red) into their respective goal regions. A circular goal region on the table is given for each object type. The goal of the robot is to place all object into the correct goal regions in the shortest amount of time.

\subsection{Sorting Algorithm}
At a given table configuration with $n$ objects, we have $2n$ high-level discrete actions. For each object, two actions are added to the action set: 

\begin{itemize}
    \itemsep0em 
    \item Picking the object and placing it to a random and collision-free position in its corresponding goal region.
    \item Pushing the object towards the center of the its goal region. Multiple objects might be pushed at once.
\end{itemize}

After each action, the robot is sent to a fixed home position. For each action, we predict the table configuration and calculate a simple heuristic where the approximated cost of a state is the sum of all object distances to their correct goal regions. We use a one-step lookahead greedy search where the action with the lowest heuristic is executed. Even though this approach is not a complete planning algorithm, it converged to sound solutions in all of our experiments.

\subsection{Methods}
We compared two methods:
\begin{itemize}
    \itemsep0em 
    \item \textbf{Pick$\&$Place}: Objects are grasped and placed to their goal regions in a random order
    \item \textbf{Push+Pick$\&$Place}: Forearm push actions are also allowed in addition to standard Pick$\&$Place actions
\end{itemize}

The total time to place all the objects to their goal regions was chosen as the operational metric.

\begin{figure}[t!]
\centering
\includegraphics[clip, trim={0cm 0cm 0cm 3cm}, width=0.25\textwidth]{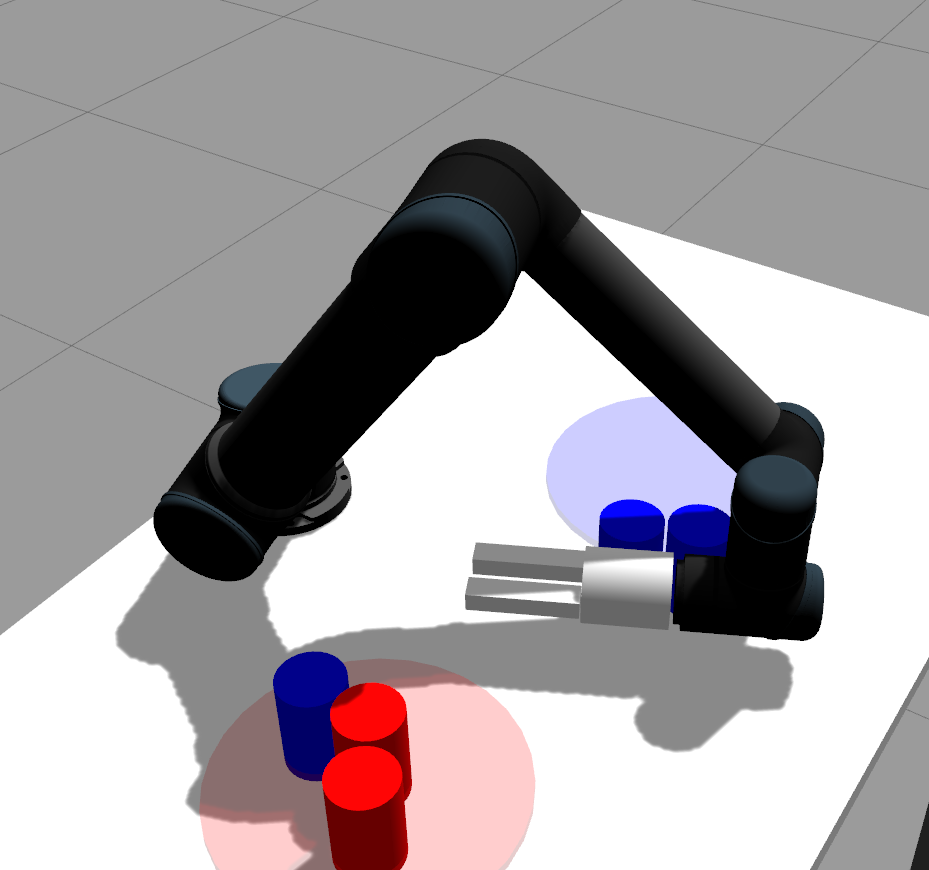}
\caption{Experiments were carried out in Gazebo simulation environment}
\vspace{-0.4cm}
\label{fig:gazebo}
\end{figure}

\subsection{Simulation Results}
We used Gazebo to simulate the sorting task, a screenshot can be seen in Fig.~\ref{fig:gazebo}. Due to difficulties we had with the gripper model, the objects were attached to the end-effector in order to simulate the grasping. The number of tabletop objects were varied from 3 to 10 where the goal regions and object positions were randomly placed at the start of each run. Both methods, were run on the same object configurations for fair comparison. Execution time from two runs for each object count are averaged to reach to the results.

\begin{figure}[t!]
\centering
\includegraphics[clip, trim={0.5cm 0.5cm 0cm 0.5cm}, width=0.45\textwidth]{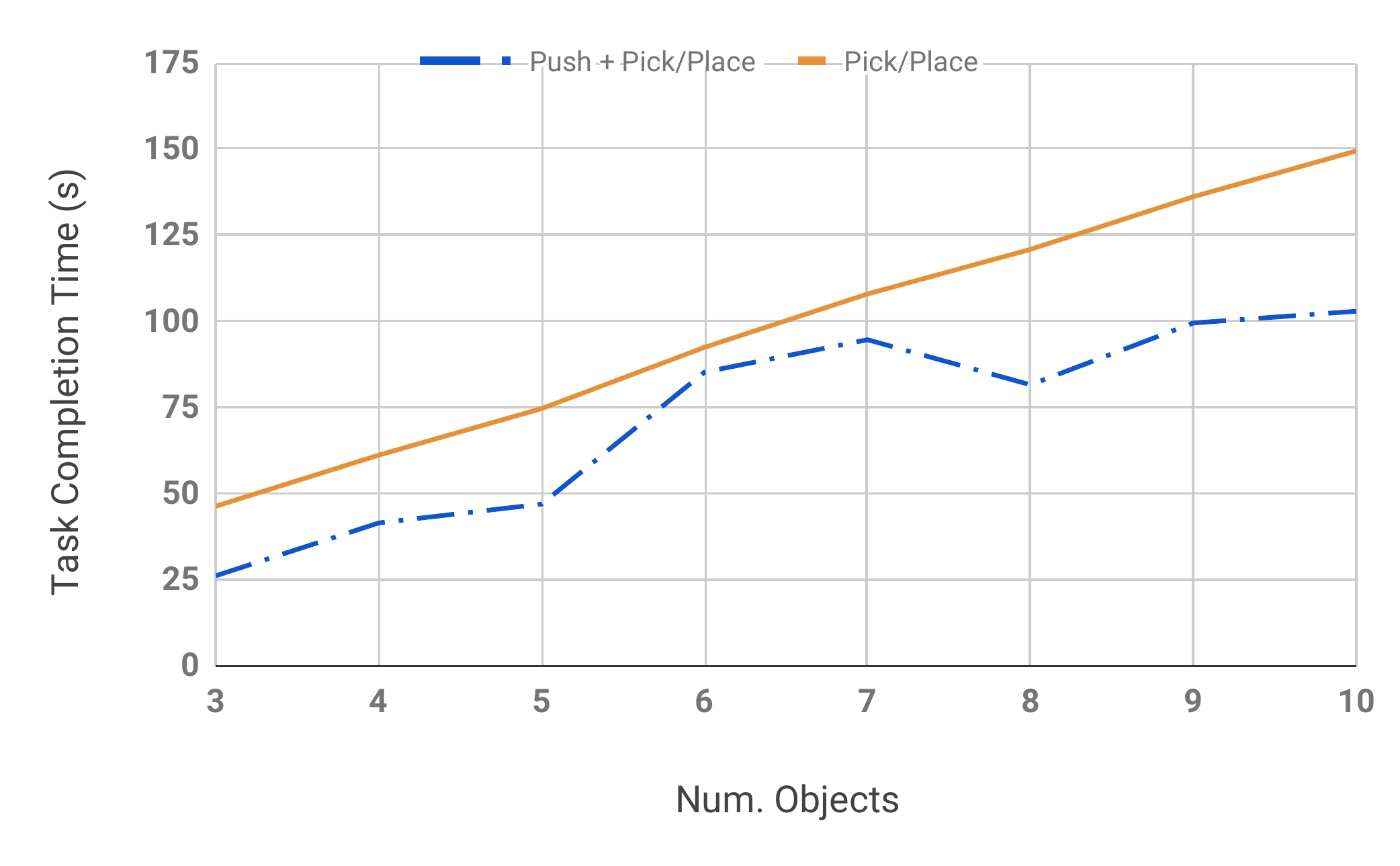}
\caption{Average task completion time is shown with respect to the number of objects on the table. The mixed strategy is consistently more efficient.}
\label{fig:plot}
\vspace{-0.4cm}
\end{figure}

The results are shown in Fig.~\ref{fig:plot}. \textbf{Pick$\&$Place} method had a linear increase in the execution time as expected, since the number of total picks is equal to the number objects. \textbf{Push + Pick$\&$Place} method was more efficient than the baseline method for all object counts and on average was \%27.9 faster in all the experiments. We observed that the greedy algorithm tends to favor big sweeping pushes at the early stages of the task and relies on pick and place at later stages.

\subsection{Robot Implementation}
The approach is implemented on a UR5 with a Robotiq gripper (Fig.~\ref{fig:intro}). Using a RGB-D camera (Kinect) across the robot, the table is detected using RANSAC and the tabletop objects are detected using simple euclidean point cloud clustering. An overhead projector is used to project the goal regions onto the table for observers to see. We selected salt objects for testing as they can be picked up easily with overhead grasps and that they don't tip over during pushing due to their weight.

\section{Conclusion and Future Work}
\label{sec:conclusion}

We propose embracing contact with the objects and demonstrate in an object sorting task that using a combination of pick$\&$place and forearm-push actions is more efficient than pick$\&$place only. Future work includes robust execution on the real robot, generalizing the approach to any of the robot's links, longer-horizon planning and reinforcement learning over the high-level action space.

\vspace{-0.1cm}
\bibliographystyle{abbrv}
{\small \bibliography{main}}

\end{document}